\documentclass[10pt,twocolumn,letterpaper]{article}

\usepackage{cvpr} % uncomment this for the final submission
\usepackage{times}
\usepackage{epsfig}
\usepackage{graphicx}
\usepackage{amsmath}
\usepackage{amssymb}
\usepackage{booktabs}
\usepackage{cuted}
\usepackage{capt-of}
\usepackage{multirow}
\usepackage{enumitem}
\newenvironment{tight_itemize}{
\begin{itemize}[leftmargin=20pt]
  \setlength{\topsep}{0pt}
  \setlength{\itemsep}{0pt}
  \setlength{\parskip}{0pt}
  \setlength{\parsep}{0pt}
}{\end{itemize}}
\begin{document}

%%%%%%%%% TITLE
\title{MoCoTalk: Multi-Conditional Diffusion with Adaptive Router for Controllable Talking Head Generation}

\author{Xinyan Ye \\
Imperial College London \\
London, UK\\
{\tt\small xinyan.ye19@imperial.ac.uk}
\and
Jiankang Deng\\
Imperial College London \\
London, UK\\
{\tt\small j.deng16@imperial.ac.uk}
\and
Abbas Edalat\\
Imperial College London \\
London, UK\\
{\tt\small a.edalat@imperial.ac.uk}
}

\maketitle
\thispagestyle{empty}

% \begin{strip}
%   \centering
%   \includegraphics[width=\linewidth]{img/diff.png}
%   \captionof{figure}{Comparison of video-only driver and audio-only driver talking head generation models with MoCoTalk. MoCoTalk provides independent controllability of facial expression, head pose, and mouth dynamics within a unified framework.}
% \end{strip}

%%%%%%%%% ABSTRACT
\begin{abstract}

Talking-head generation requires joint modeling of identity, head pose,
facial expression, and mouth dynamics. Existing methods typically address only a subset of these factors, and rely on fixed-weight or heuristic fusion when multiple conditions are involved. We present MoCoTalk, a multi-conditional video diffusion framework that unifies four complementary control signals: a reference image, facial keypoints, 3DMM-rendered shading meshes, and the corresponding speech audio. To resolve destructive interference among heterogeneous conditions, we introduce an Adaptive Multi-Condition Router that computes channel-wise, timestep-aware gating over the four condition streams, allowing the fusion strategy to vary with both feature subspace and noise level. To better capture speech-related facial dynamics, we design a Mouth-Augmented Shading Mesh, a 3DMM-based representation that decouples head motion, mouth motion, expression, and lighting. This design provides a temporally consistent geometric prior and allows flexible recombination of these attributes at inference. We further introduce a lip consistency loss to tighten audio-visual alignment. Extensive experiments show that MoCoTalk achieves state-of-the-art performance on the majority of structural, motion, and perceptual metrics, while offering attribute-level controllability that single-condition methods do not provide.

\end{abstract}

%%%%%%%%% BODY TEXT
\section{Introduction}
Talking head generation aims to animate a static portrait image according to specific driving conditions, such as motion sequences or speech signals. The objective is to generate photorealistic videos in which the animated portrait preserves the source identity while exhibiting natural, audio-visually aligned dynamics. This task underpins a wide range of applications, including virtual avatars, video conferencing, digital dubbing, and interactive entertainment~\cite{meng2026surveytalkinghead,xu2023progressivetransformermachine,liu2023talkingfacegeneration}.

Depending on the modality of the driving input, existing approaches can be broadly categorized into video-driven and audio-driven methods. Video-driven approaches transfer motion from a driving video to a target identity, typically leveraging structured representations such as facial landmarks, segmentation maps~\cite{xiong2024segtalkersegmentationbasedtalking}, or rendered 3D meshes~\cite{guo2025highfidelityrelightablemonocular}, whose expressiveness is ultimately bounded by the underlying tracking accuracy. In contrast, audio-driven methods focus on synchronizing lip movements with speech signals by mapping audio features into the visual domain. However, this cross-modal mapping is inherently one-to-many, as the same audio can correspond to multiple plausible facial motions, making controllability a critical concern.

\begin{figure}[t]
\centering
\includegraphics[width=0.98\linewidth]{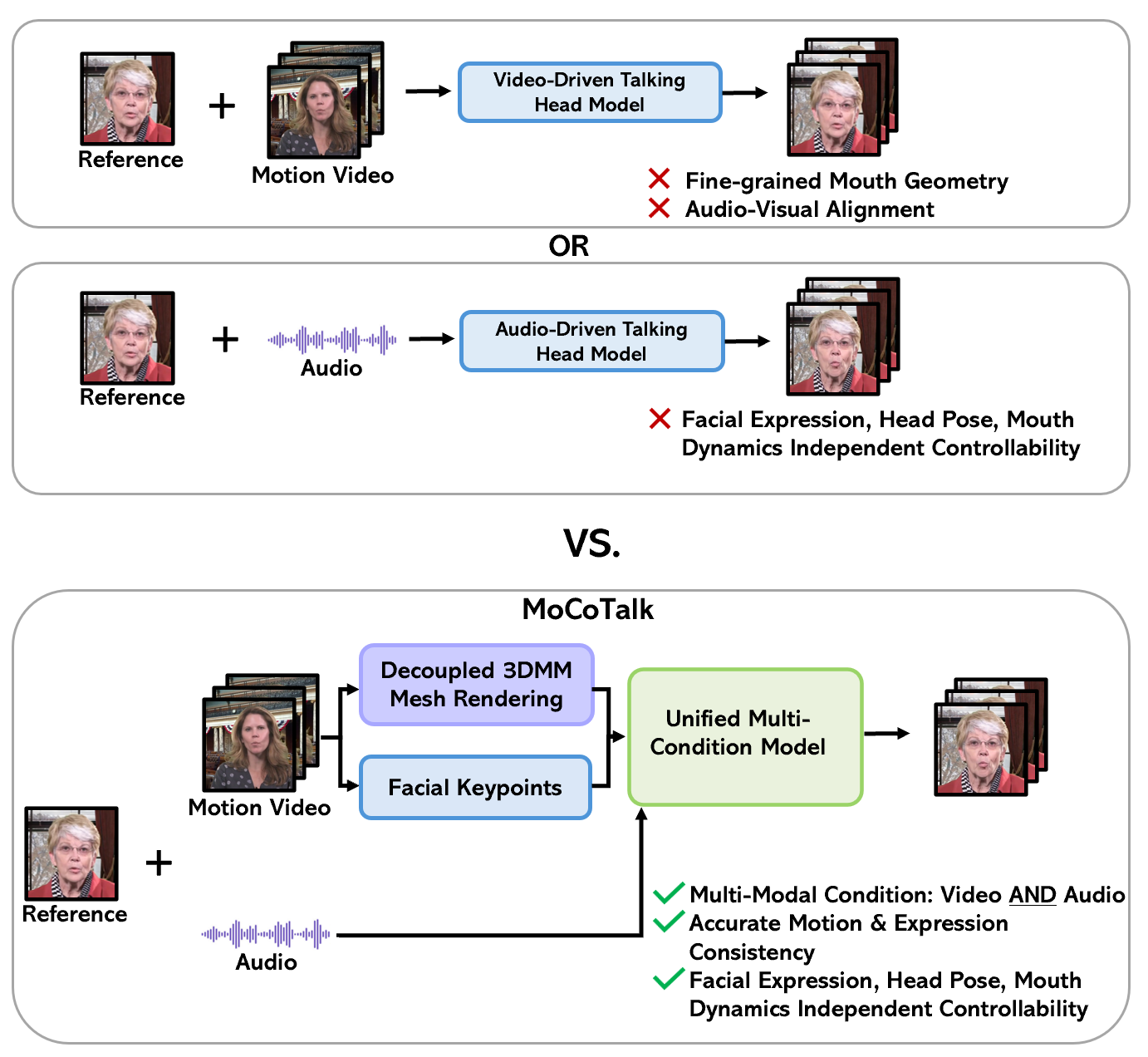}
\caption{Comparison of video-only driver and audio-only driver talking head generation models with MoCoTalk. MoCoTalk provides independent controllability of facial expression, head pose, and mouth dynamics within a unified framework.}
\label{fig:diff}
\end{figure}

Facial dynamics involve tightly coupled factors, including head pose, facial expressions, and lip movements, which are globally coordinated in natural human behavior. Disentangling and independently controlling these attributes is therefore non-trivial. Prior works attempt to address this challenge through explicit decomposition or latent disentanglement. Liang \etal~\cite{liang2022expressivetalkinghead} address this by decomposing the reference image into a cropped mouth region, a masked head, and the upper face. Xiong \etal ~\cite{xiong2024segtalkersegmentationbasedtalking} decouple lip motion from portrait image using facial segmentation masks as an intermediate representation. Other works disentangle the latent space into factors corresponding to mouth, pose, and expression, allowing targeted manipulation without disturbing unrelated visual components~\cite{wang2022latentimageanimator,meshry2021learnedspatialrepresentations,zakharov2020fastbilayerneural}. While these approaches improve controllability and enable targeted editing, they often struggle to maintain global coherence, especially under cross-reenactment settings where motion from one identity is transferred to another. Common failure cases include drifting facial features and mismatched mouth-region geometry.

With the rapid advancement of denoising diffusion probabilistic models, generative modeling has achieved significant improvements in visual quality and diversity, consistently surpassing GAN-based approaches~\cite{karras2019stylebasedgeneratorarchitecture}. In particular, latent diffusion models~\cite{song2022denoisingdiffusionimplicit} enable iterative refinement of latent representations under external conditioning signals. Recent works~\cite{zhang2023addingconditionalcontrol,mou2023t2iadapterlearningadapters} demonstrate the effectiveness of incorporating spatial and semantic conditions into diffusion frameworks. Extending this to talking head generation, however, raises questions that go beyond single-condition generation: the joint modeling of head pose, lip synchronization, facial expression and reference identity within a unified framework remains largely open. In particular, it is unclear how to balance the contributions of diverse conditioning signals, and how to fuse cross-modal information in a coherent and adaptive manner.

In this paper, we present \textbf{MoCoTalk}, a novel multi-conditional diffusion framework that accepts four control signals: audio, facial keypoints, 3D mesh, and a reference image, covering the complementary aspects of motion, geometry, and appearance. Unlike existing approaches~\cite{guo2025highfidelityrelightablemonocular} that rely on fixed-weight fusion, we introduce an Adaptive Multi-Condition Router, a lightweight adaptive gating module that dynamically weights and combines heterogeneous inputs in the latent space. To strengthen motion modeling, we build on a 3DMM-based representation and propose a fusion mechanism that jointly captures fine-grained mouth articulation and global facial structure.

In summary, our contributions are as follows:
\begin{tight_itemize}
    \item We propose a novel multi-conditional diffusion framework
    for talking-head video generation that integrates four
    heterogeneous conditioning signals: audio, keypoints, 3D mesh,
    and a reference image.
    \item We introduce an \textbf{Adaptive Multi-Condition Router}, a lightweight
    adaptive gating module that dynamically weights and combines
    heterogeneous control signals in the latent space.
    \item We design a \textbf{Mouth-Augmented Mesh Fusion} that
    decouples and recombines motion sources to render 3DMM
    meshes that are tightly aligned with speaking dynamics, providing a
    temporally consistent geometric prior.
    \item We further introduce a \textbf{Lip Consistency
    Loss} that supervises mouth-region semantics in pixel space,
    improving speech-related lip dynamics.
\end{tight_itemize}

%-------------------------------------------------------------------------
% ============================================================
%  Method section
%  Requires: \usepackage{amsmath, amssymb}
%  Optional: \usepackage{bm} for bold math
% ============================================================

\section{Related Work}
\label{sec:related work}

\subsection{Implicit and Explicit Facial Representations}

Approaches to modeling facial motion can be divided into those based on implicit and explicit representations.

\noindent \textbf{Implicit Facial Representations.} Implicit representation methods encode facial motion in a learned latent space, typically through implicit keypoints or motion fields that are transferred from a driving source to a target portrait. First Order Motion Model (FOMM)~\cite{siarohin2020firstordermotion} pioneered this direction by encoding motion as a set of self-supervised learned keypoints together with their local affine transformations. LivePortrait~\cite{guo2025liveportraitefficientportrait} achieves high-fidelity animation by learning compact implicit motion representations that support stitching and retargeting control, demonstrating that carefully scaled implicit approaches remain highly competitive. EmoPortrait~\cite{drobyshev2024emoportraitsemotionenhancedmultimodal} incorporates emotional state conditioning into the latent space, enabling expression-aware animation from a reference image. Other methods adapt to different facial dynamics and controls by disentangling the latent spaces to represent the mouth, pose, and expression~\cite{wang2022latentimageanimator,meshry2021learnedspatialrepresentations,zakharov2020fastbilayerneural}. However, these works often display significant artifacts when there is a large disparity between the driving image and the reference image. Moreover, the lack of interpretable intermediate representations limits their ability to independently manipulate attributes such as lighting and fine-grained lip articulation, which is critical for precise and controllable animation.

\noindent \textbf{Explicit Facial Representations.} Explicit methods instead decouple facial motion into semantically meaningful signals. Facial landmark sequences provide a sparse but geometrically grounded representation of pose and expression, and have been adopted as driving signals in a range of work~\cite{wei2024aniportraitaudiodrivensynthesis,chen2024echomimiclifelikeaudiodriven,ma2024followyouremojifinecontrollableexpressive}. While effective for coarse motion transfer, landmarks are inherently limited in capturing detailed facial geometry and subtle articulations. More expressive control is achieved through parametric models such as 3D Morphable Models (3DMMs)~\cite{blanz1999morphablemodelsynthesis} and their parametric successors~\cite{li2017learningmodelfacial,feng2021learninganimatabledetailed} decompose faces into disentangled coefficients representing shape, pose, expression, and illumination. Several works~\cite{doukas2021headganoneshotneural,ren2021pirenderercontrollableportrait,yin2022styleheatoneshothighresolution} exemplify this paradigm by explicitly decoupling pose and expression to enable targeted editing. More recently, LCVD~\cite{guo2025highfidelityrelightablemonocular} leverages a static 3D face reconstruction model, DECA~\cite{feng2021learninganimatabledetailed}, to render a shaded, motion-aware mesh, enabling relighting control over the generated portrait at inference. Despite their parametric controllability, 3DMM-based representations are fundamentally constrained by the expressiveness of the parametric model and the accuracy of the fitting process~\cite{egger20203dmorphableface}. In particular, they often underfit fine-grained mouth-region dynamics, failing to capture subtle lip articulations and jaw movements, which leads to a structural mismatch between the mesh geometry and the actual lip motion. To address this limitation, we incorporate a pretrained model, SPECTRE~\cite{filntisis2022visualspeechawareperceptuala}, that recovers temporally consistent lip geometry from video, and design a mouth-augmented mesh representation that enhances the capture of mouth-region dynamics while preserving the global controllability offered by parametric models.

\subsection{Diffusion-based Talking Head Generation}

The emergence of large-scale denoising diffusion probabilistic models~\cite{ho2020denoisingdiffusionprobabilistic} and their  latent variants~\cite{song2022denoisingdiffusionimplicit} has substantially raised the quality ceiling for talking head generation. Early diffusion-based approaches adapt pretrained text-to-image models such as Stable Diffusion~\cite{rombach2022highresolutionimagesynthesis} by introducing additional conditioning mechanisms. For example, AnimateDiff~\cite{guo2024animatediffanimateyour} incorporates learnable temporal attention layers into the U-Net between frozen spatial layers to learn motion priors, while DiffTalk~\cite{shen2023difftalkcraftingdiffusion} and SadTalker~\cite{zhang2023sadtalkerlearningrealistic} inject audio features and 3DMM coefficients via cross-attention to enable speech-driven facial animation. Subsequent work has explored identity-preserving generation, typically by introducing auxiliary appearance branches that extract dense identity
features from the reference image and inject them into the denoising
U-Net~\cite{hu2024animateanyoneconsistent,tu2024stableanimatorhighqualityidentitypreserving}. A more recent line of work migrates to Stable Video Diffusion (SVD)~\cite{blattmann2023stablevideodiffusion} as the generative backbone in order to exploit its stronger spatio-temporal priors over natural video dynamics. Champ~\cite{zhu2024champcontrollableconsistent}, Follow-Your-Emoji~\cite{ma2024followyouremojifinecontrollableexpressive}, and X-Portrait~\cite{xie2024xportraitexpressiveportrait} build end-to-end human video generation pipelines, conditioning SVD on various combinations of rendered 3D geometry, landmark sequences, and reference appearance. Beyond motion and identity control, LCVD~\cite{guo2025highfidelityrelightablemonocular} extends SVD-based portrait animation to the relightable setting by decomposing intrinsic and extrinsic feature subspaces via a shading adapter and a reference adapter, thereby enabling independent control over lighting and identity.
Despite these advances, existing approaches typically address only a subset of control factors, and often rely on fixed-weight or heuristic fusion strategies when multiple conditions are involved. In this work, we address this limitation by introducing a multi-conditional diffusion architecture with four dedicated adapters for four complementary signals (audio, keypoints, mesh, and reference), and by fusing them inside a single denoising U-Net through an adaptive gating mechanism.

\section{Preliminaries}
\label{sec:preliminaries}

Stable Video Diffusion (SVD)~\cite{blattmann2023stablevideodiffusion} extends the Latent Diffusion Model (LDM)~\cite{rombach2022highresolutionimagesynthesis} to image-to-video generation by introducing temporal layers into the 2D U-Net, enabling joint spatio-temporal modeling within the latent space of a Variational Autoencoder (VAE)~\cite{esser2021tamingtransformershighresolution}. Given a video clip $\mathbf{x}_0 \in \mathbb{R}^{N \times 3 \times H \times W}$, the VAE encoder $\mathcal{E}(\cdot)$ maps it to the latent $\mathbf{z}_0 = \mathcal{E}(\mathbf{x}_0)$, which is perturbed via a Markov chain:
\begin{equation}
    \mathbf{z}_t = \sqrt{\bar{\alpha}_t}\, \mathbf{z}_0 + \sqrt{1 - \bar{\alpha}_t}\, \boldsymbol{\epsilon}, \quad \boldsymbol{\epsilon} \sim \mathcal{N}(\mathbf{0}, \mathbf{I}),
\end{equation}
where $\bar{\alpha}_t = \prod_{i=1}^{t}(1 - \beta_i)$ and $t \sim \mathcal{U}\{1, \ldots, T\}$. Conditioned on a reference image $\mathbf{c}_I$ injected via both CLIP~\cite{radford2021learningtransferablevisualb} cross-attention and channel-wise concatenation with $\mathbf{z}_t$, the denoising network $\boldsymbol{\epsilon}_\theta$ is trained with:
\begin{equation}
    \mathcal{L}_{\text{SVD}} = \mathbb{E}_{\mathbf{z}_0, \mathbf{c}_I, \boldsymbol{\epsilon}, t} \left[ \left\| \boldsymbol{\epsilon} - \boldsymbol{\epsilon}_\theta(\mathbf{z}_t, t, \mathbf{c}_I) \right\|_2^2 \right].
\end{equation}

\begin{figure*}
\begin{center}
% \fbox{\rule{0pt}{2in} \rule{.9\linewidth}{0pt}}
   \includegraphics[width=1\linewidth]{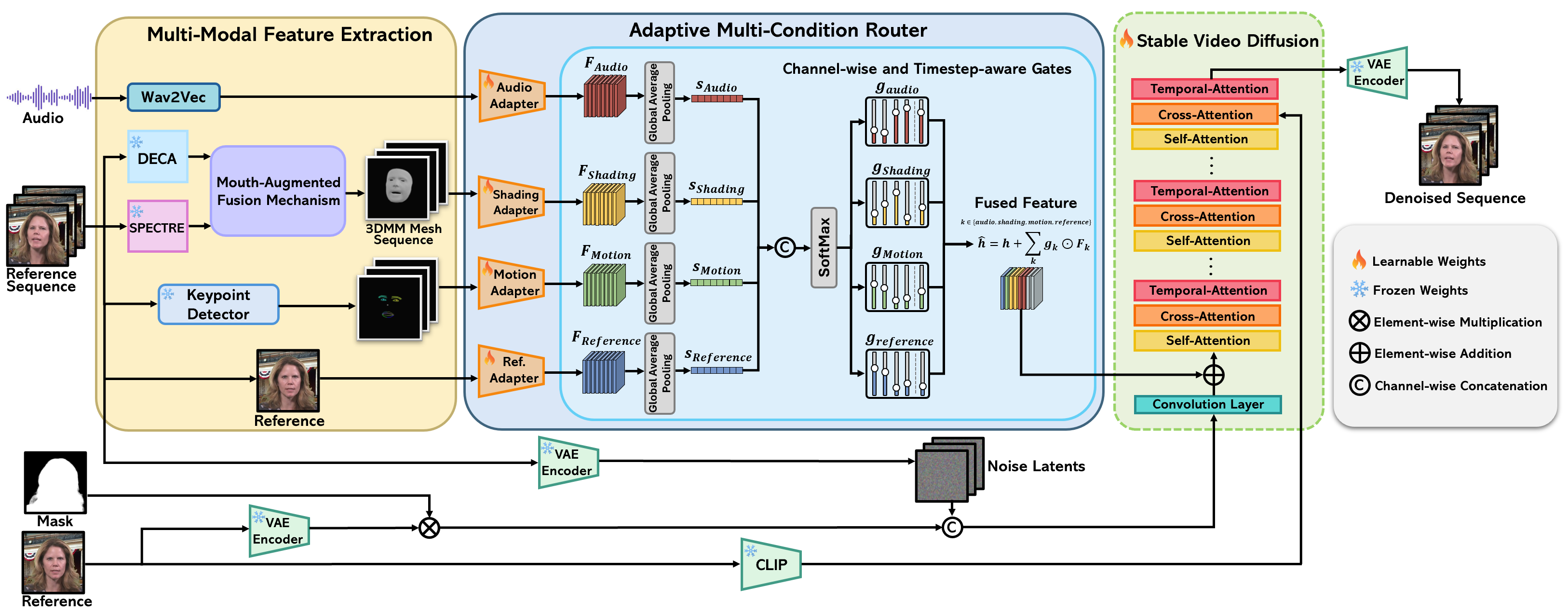}
\end{center}
   \caption{\textbf{Overview of the multi-conditional video diffusion 
framework.} 
MoCoTalk accepts four complementary conditioning signals: a reference 
portrait, facial keypoints, mouth-augmented 3DMM shading meshes, and a speech audio. Lighting, shape, pose, and expression parameters are 
extracted from video frames using DECA~\cite{feng2021learninganimatabledetailed} 
and SPECTRE~\cite{filntisis2022visualspeechawareperceptuala}, and fused via our four-source pipeline to render the 
mouth-augmented shading meshes (Sec.~\ref{sec:method_shading}). The audio track is encoded by 
Wav2Vec~\cite{schneider2019wav2vecunsupervisedpretraining} into a windowed feature sequence. 
The four signals are lifted into spatially aligned latent feature maps 
by four parallel adapters, and then fed into the Adaptive Multi-Condition Router (Sec.~\ref{sec:method_router}), which adaptively fuses the heterogeneous 
signals into the shallow U-Net feature via channel-wise gating. The fused feature 
guides a Stable Video Diffusion 
backbone~\cite{blattmann2023stablevideodiffusion} to denoise the video latents, yielding generated sequence with consistent identity, lighting, head 
pose, facial expression and mouth motion.}
\label{fig:train}
\end{figure*}

\section{Method}
\label{sec:method}

\subsection{Overview}
\label{sec:method_overview}

We formulate talking-head generation as a conditional video denoising problem built upon SVD~\cite{blattmann2023stablevideodiffusion}.
Given a reference portrait, a target head motion sequence, and a driving speech track, our model generates a video that preserves the reference identity and lighting while following the specified head pose and lip articulation. To bridge raw inputs and the latent denoiser,  we further derive two intermediate visual controls: facial keypoint maps representing pose and coarse expression, and 3DMM-rendered shading meshes representing geometry and lighting.

As illustrated in Figure~\ref{fig:train}, our framework consists of four components: (i) four parallel adapters
that lift each condition into a spatially aligned latent
(Sec.~\ref{sec:method_encoders}); (ii) an Adaptive Multi-Condition Router that performs channel-wise, timestep-aware fusion
(Sec.~\ref{sec:method_router}); (iii) a Mouth-Augmented Mesh Rendering pipeline that resolves the lower-face artefacts of vanilla DECA
(Sec.~\ref{sec:method_shading}); and (iv) a Lip Consistency Loss that tightens audio-visual alignment (Sec.~\ref{sec:method_lip}).

% ------------------------------------------------------------
\subsection{Multi-Conditional Adapters}
\label{sec:method_encoders}

We introduce four parallel adapters into the SVD backbone, each
mapping a conditioning modality into a latent feature
map that is spatially and channel-wise aligned with the shallow U-Net
features:

\begin{tight_itemize}
    \item \textbf{Reference Adapter} $\mathcal{F}_r$ takes the
    reference portrait as input and extracts identity and appearance
    features.
    \item \textbf{Motion Adapter} $\mathcal{F}_m$ takes a sequence
    of keypoint maps as input and encodes head pose and coarse
    expression.
    \item \textbf{Shading Adapter} $\mathcal{F}_s$ takes a sequence
    of 3DMM-based mesh renderings as input and captures facial
    geometry and lighting.
    \item \textbf{Audio Adapter} $\mathcal{F}_a$ takes a windowed speech signal as input and extracts audio features.
\end{tight_itemize}

\noindent{ \bf Spatial audio projection.}
Unlike prior work that injects audio as tokens through cross-attention, we project audio into a spatial latent feature map
that is spatially compatible with the three visual branches. The audio stream can thus participate in the same feature-level fusion as the visual controls, without requiring an additional token-level attention module.

For each target frame
$t$, we form a local temporal window of length $W = 2m+1$ ($m=2$),
\begin{equation}
    \mathbf{A}_t = \{\, a_{t-m}, \ldots, a_t, \ldots, a_{t+m} \,\}.
\end{equation}
Stacking across $T$ frames yields
$\mathbf{A} \in \mathbb{R}^{T \times W \times L \times C_a}$, where
$L$ and $C_a$ denote token length and feature dimensionality. The audio adapter then maps $\mathbf{A}$ to spatial latents
$\mathbf{F}_{\text{audio}} = \mathcal{F}_a(\mathbf{A})$, which are gated and
fused with the three visual streams in Sec.~\ref{sec:method_router}.
The sliding window preserves local temporal context and exposes audio to the same fusion pathway as the visual conditions.

% ------------------------------------------------------------
\subsection{Adaptive Multi-Condition Router}
\label{sec:method_router}

Existing multi-condition designs typically assign each
condition an independent scalar weight, which makes it difficult to
adaptively decide which modality should dominate under different
frames and noise levels. We propose a lightweight router immediately after the input convolution of the U-Net, which dynamically balances the four
heterogeneous conditions by calculating channel-wise gates conditioned on both the backbone feature and all available condition signals.

Let $\mathbf{h} \in \mathbb{R}^{BT \times C \times h \times w}$ denote the shallow U-Net feature after the input convolution, where $B$ is the batch size and $T$ is the number of frames. For each condition branch $k \in \{\text{reference}, \text{shading}, \text{motion}, \text{audio}\}$, we compute a global summary by spatial average pooling:
\begin{equation}
\mathbf{s}_k = \mathrm{GAP}(\mathbf{F}_k), \quad
\mathbf{u} = \mathrm{GAP}(\mathbf{h}), \quad
\mathbf{t} = \phi(\mathbf{e}),
\end{equation}

% \begin{align}
% \mathbf{s}_k &= \mathrm{GAP}(\mathbf{F}_k), \\
% \mathbf{u} &= \mathrm{GAP}(\mathbf{h}), \\
% \mathbf{t} &= \phi(\mathbf{e}),
% \end{align}

where $\mathbf{e}$ is the timestep embedding, 
and $\phi(\cdot)$ is a linear projection to channel dimension $C$.

The router concatenates these summaries and feeds them through a 
lightweight MLP $\psi(\cdot)$, whose output is normalized by a softmax across the four condition branches:
\begin{equation}
\mathbf{g}
=
\mathrm{Softmax}\!\left(
\psi\!\left([\mathbf{u}, \mathbf{t}, \mathbf{s}_{\text{reference}}, 
\mathbf{s}_{\text{shading}}, \mathbf{s}_{\text{motion}}, 
\mathbf{s}_{\text{audio}}]\right)
\right),
\end{equation}
% yielding gates $\mathbf{g} \in \mathbb{R}^{BT \times 4 \times C}$ 
% that sum to one per channel.

The fused feature is then computed as
\begin{equation}
\tilde{\mathbf{h}}
=
\mathbf{h}
+ \sum_{k} \mathbf{g}_{k} \odot \mathbf{F}_{k}
\end{equation}

where each gate $\mathbf{g}_k \in \mathbb{R}^{BT \times C \times 1 \times 1}$ is channel-wise and $\odot$ denotes broadcast multiplication. For conditions dropped during training or unavailable at inference,
the corresponding logits are masked to a large negative constant, which effectively zeroes their contribution after softmax. This formulation has three properties tailored to multi-conditional
talking-head synthesis. Channel-wise gating allocates distinct
feature subspaces to different modalities rather than collapsing them
into a single scalar weight. Timestep-aware fusion lets the
router shift its emphasis along the denoising trajectory. Mask-based dropout makes the router robust to missing
modalities, which we leverage for both random condition dropout in training and partial-control inference.

% ------------------------------------------------------------
\subsection{Mouth-Augmented Shading Mesh}
\label{sec:method_shading}
DECA~\cite{feng2021learninganimatabledetailed} is optimized for
global identity and expression recovery from static images. However, it
consistently under-fits the mouth region. When applied frame-by-frame to a speech video, the resulting mesh sequence tends to smear fine-grained lip articulation and exhibits temporal jitter. Both artefacts propagate to the lower-face appearance flow. We address this with a four-source fusion pipeline that preserves DECA's identity and coarse-expression priors while injecting SPECTRE's~\cite{filntisis2022visualspeechawareperceptuala} mouth tracking.

Given a target speech video $\mathcal{V}_{\text{speech}}$, an optional
head-motion video $\mathcal{V}_{\text{head}}$, a reference frame
$I_{\text{ref}}$, and a lighting reference frame $I_{\text{light}}$,
the pipeline produces a per-frame shading map
$\mathcal{S}_t \in \mathbb{R}^{512 \times 512}$ in four steps.

\noindent{\bf Step 1. Identity and lighting extraction.}
We fit DECA to $I_{\text{ref}}$ to obtain the identity shape
$\mathbf{s} \in \mathbb{R}^{100}$ and camera parameters
$\mathbf{c} \in \mathbb{R}^{3}$, and to $I_{\text{light}}$ to obtain
the 27-dimensional spherical-harmonic illumination
$\boldsymbol{\ell} \in \mathbb{R}^{27}$ (9 SH coefficients $\times$ 3 channels). These parameters are held constant across
all output frames.

\noindent{\bf Step 2. Mouth-aware motion tracking.}
For each input video, SPECTRE predicts a base FLAME expression
$\mathbf{e}_{\text{SPECTRE}} \in \mathbb{R}^{50}$ together with a
residual DECA correction
$\Delta\mathbf{e}_{\text{DECA}} \in \mathbb{R}^{50}$, and analogously
for jaw pose.
\begin{equation}
    \mathbf{e}_{\text{exp}}
        = \mathbf{e}_{\text{SPECTRE}} + \Delta\mathbf{e}_{\text{DECA}},
    \qquad
    \boldsymbol{\theta}_{\text{jaw}}
        = \boldsymbol{\theta}_{\text{SPECTRE}}
          + \Delta\boldsymbol{\theta}_{\text{DECA}}.
\end{equation}

\noindent{\bf Step 3. Cross-source motion fusion.}
At frame $t$, we take the global head rotation
$\mathbf{r}_t^{\text{head}} \in \mathbb{R}^{3}$ from
$\mathcal{V}_{\text{head}}$, and the jaw pose
$\boldsymbol{\theta}_t^{\text{jaw}} \in \mathbb{R}^{3}$ together with
the facial expression $\mathbf{e}_t^{\text{exp}} \in \mathbb{R}^{50}$
from $\mathcal{V}_{\text{speech}}$. The fused pose used
for rendering is
\begin{equation}
    \boldsymbol{\theta}_t^{\text{pose}}
        = [\,\mathbf{r}_t^{\text{head}},\;
             \boldsymbol{\theta}_t^{\text{jaw}}\,].
\end{equation}
This decoupling allows head orientation and
speech-driven mouth dynamics to be sourced from independent videos.

\noindent{\bf Step 4. Shading rendering.}
Using DECA's differentiable renderer, we obtain the per-frame shading map that compactly encodes 3D geometry, pose, and
illumination. 
\begin{equation}
    \mathcal{S}_t = \mathrm{Render}(
        \mathbf{s},\,
        \mathbf{e}_t^{\text{exp}},\,
        \boldsymbol{\theta}_t^{\text{pose}},\,
        \mathbf{c},\,
        \boldsymbol{\ell}
    ).
\end{equation}
The resulting sequence $\{\mathcal{S}_t\}$ serves as the
conditioning input to the shading encoder $\mathcal{F}_s$.

\begin{figure}[t]
\begin{center}
   \includegraphics[width=1\linewidth]{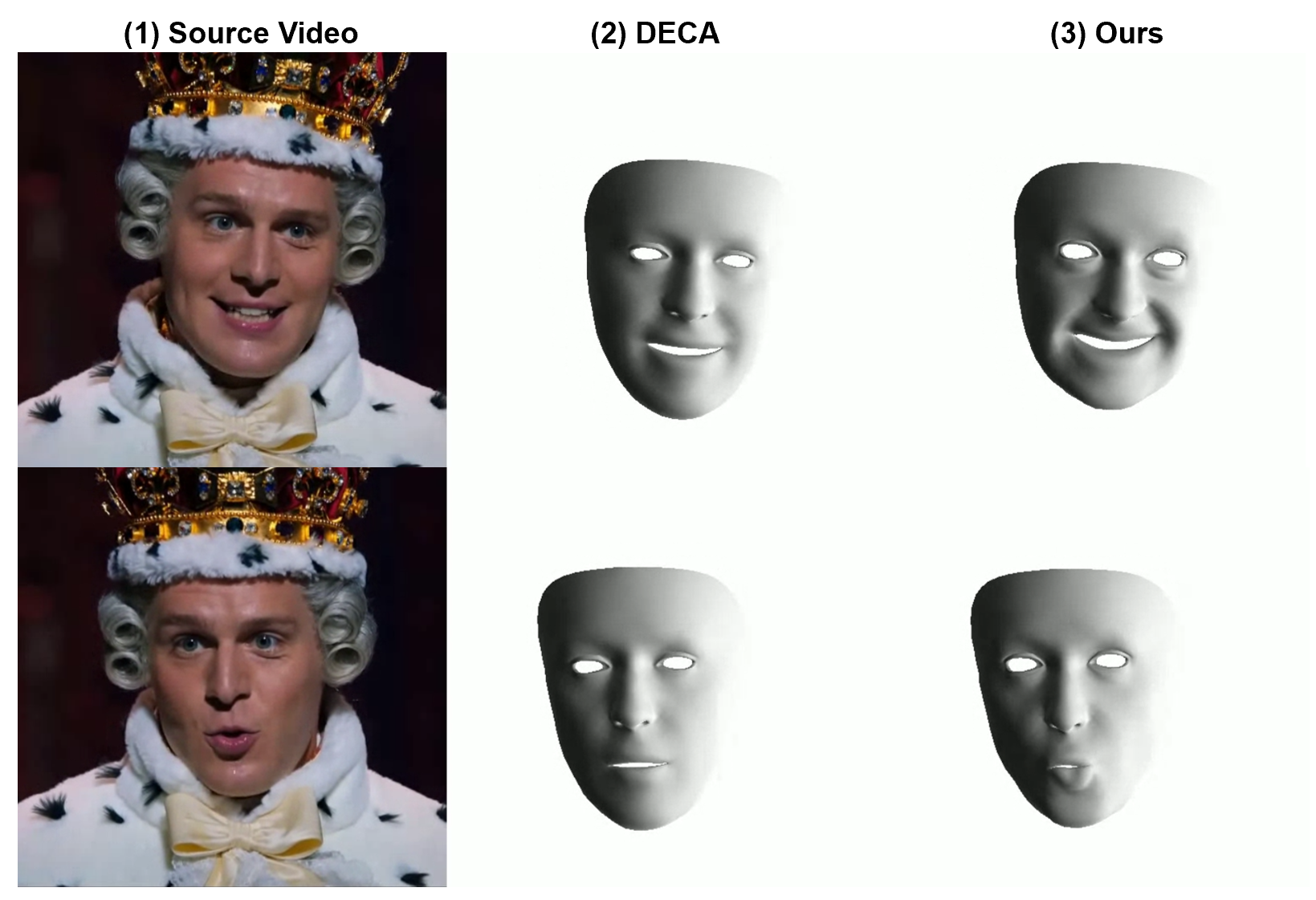}
\end{center}
   \caption{Comparison of 3DMM mesh rendering results. (1) Source video; (2) DECA-only; (3) Our mouth-augmented fusion. A right-to-left lighting is added at render time to enhance visual contrast and highlight geometric details.
   }
\label{fig:mouth}
\end{figure}

\begin{figure}[t]
\begin{center}
   \includegraphics[width=1\linewidth]{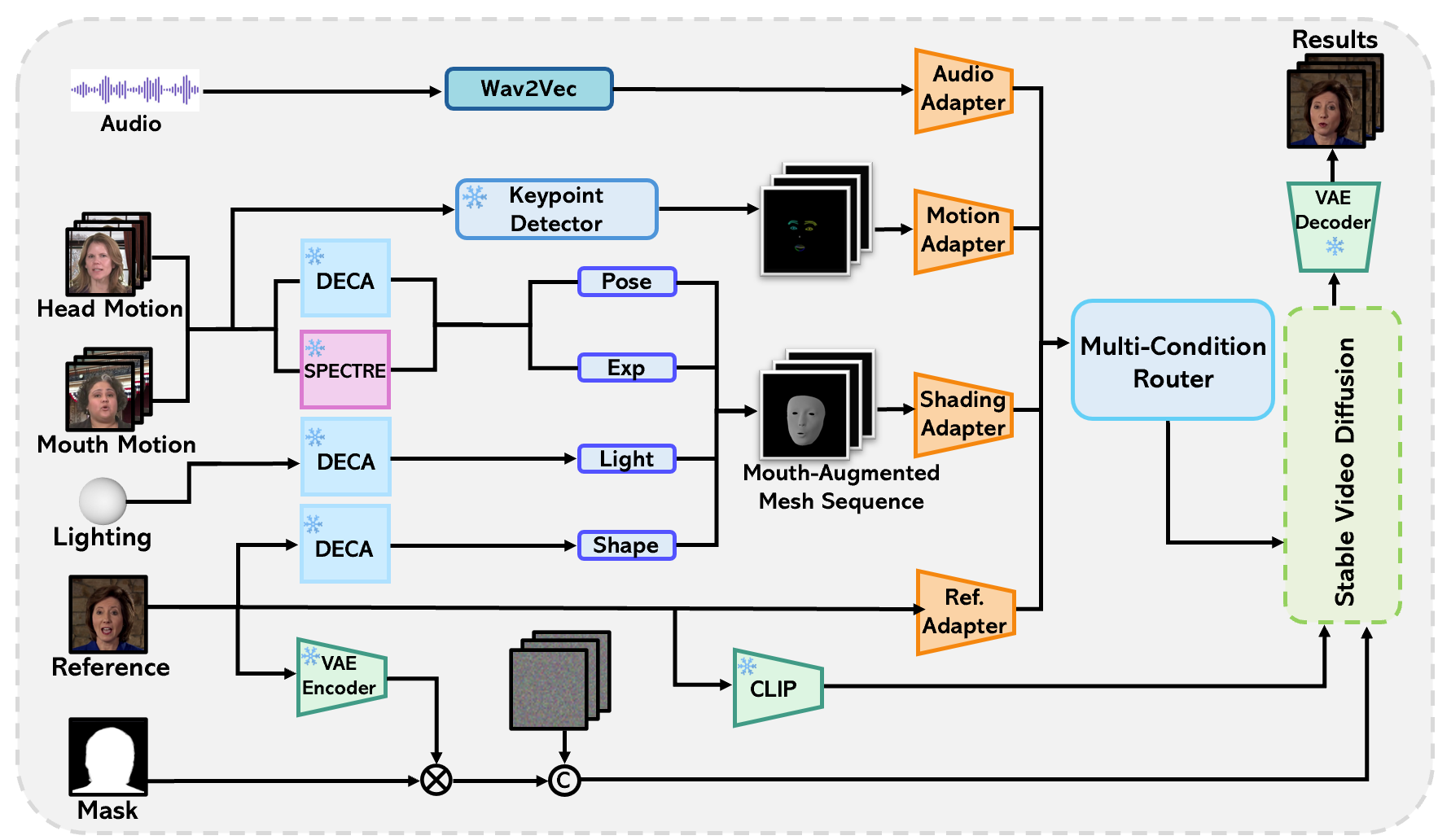}
\end{center}
   \caption{\textbf{Attribute-level Controllability of MoCoTalk.} The four-source fusion design decouples identity, lighting, head motion, and mouth motion, allowing each attribute to be drawn from an independent source and freely recombined at inference.}
   \label{fig:inference}
\end{figure}

By explicitly augmenting the rendering pipeline with mouth-aware tracking cues, the resulting mesh sequence provides a richer appearance-flow signal around the lower face, as shown in Figure~\ref{fig:mouth}. Moreover, our fusion design enables a form of controllability not available in previous pipelines: because identity, lighting, head motion, and mouth dynamics are each drawn from an independent source, they can be freely recombined at inference, illustrated in Figure~\ref{fig:inference}. 
For example, a neutral frontal head motion
can be replaced with an expressive nodding sequence while preserving the original lip dynamics. Given a specified identity, head motion, mouth motion, and the corresponding audio, this yields a customizable and audio-visually aligned mesh rendering.

\begin{table*}[t]
\centering
\resizebox{\textwidth}{!}{
\begin{tabular}{|l|c|c|c|c|c|c|c|c|c|c|c|c|c|}
\hline
\multirow{2}{*}{Method} & \multicolumn{10}{c|}{Self-Reenactment} & \multicolumn{3}{c|}{Cross-Reenactment} \\
\cline{2-14}
 & SSIM$\uparrow$ & PSNR$\uparrow$ & LPIPS$\downarrow$ & FVD$\downarrow$ & AED$\downarrow$ & APD$\downarrow$ & AKD$\downarrow$ & VQA$\uparrow$ & ID$\uparrow$ & Sync-C$\uparrow$ & AED$\downarrow$ & APD$\downarrow$ & ID$\uparrow$ \\
\hline\hline
VideoReTalking \cite{cheng2022videoretalkingaudiobasedlip} & 0.7365 & 22.09 & 0.0976 & 279.9 & 0.1984 & 50.38 & 3.239 & 0.5526 & 0.6039 & 4.153 & 0.2387 & 90.72 & 0.8488 \\
Diff2Lip \cite{mukhopadhyay2024diff2lipaudioconditioned} & \underline{0.7415} & \underline{22.55} & \underline{0.0916} & 211.3 & 0.1277 & 46.33 & 2.766 & 0.5596 & 0.8713 & \textbf{4.479} & 0.2471 & 90.86 & 0.8703 \\
SadTalker \cite{zhang2023sadtalkerlearningrealistic} & 0.7290 & 22.02 & 0.1080 & \underline{64.66} & \underline{0.1162} & 54.66 & 3.060 & 0.4691 & \textbf{0.9169} & 3.969 & 0.2384 & 95.46 & \textbf{0.9159} \\
FLOAT \cite{ki2025floatgenerativemotion} & 0.4515 & 14.58 & 0.3092 & 91.88 & 0.1360 & 76.11 & 3.946 & \underline{0.7192} & 0.8857 & 4.334 & \textbf{0.2348} & 90.09 & 0.8815 \\
Hallo2 \cite{cui2024hallo2longdurationhighresolution} & 0.7152 & 21.78 & 0.0983 & 87.62 & 0.1205 & 44.72 & 2.708 & 0.6873 & \underline{0.8929} & \underline{4.443} & 0.2434 & \textbf{86.66} & \underline{0.8923} \\
LCVD \cite{guo2025highfidelityrelightablemonocular} & 0.7209 & 21.48 & 0.1093 & 97.45 & 0.1234 & \underline{13.90} & \underline{1.529} & \textbf{0.7477} & 0.7762 & 1.163 & 0.2466 & 91.26 & 0.7705 \\
\hline
Ours & \textbf{0.7998} & \textbf{23.55} & \textbf{0.0810} & \textbf{40.75} & \textbf{0.1028} & \textbf{13.85} & \textbf{1.393} & 0.6320 & 0.8382 & 1.191 & \underline{0.2382} & \underline{89.60} & 0.8293 \\
\hline
\end{tabular}
}
\vspace{0.005pt} 
\caption{Quantitative results of talking-head generation under self-reenactment and cross-reenactment settings. We compare against VideoReTalking~\cite{cheng2022videoretalkingaudiobasedlip}, Diff2Lip~\cite{mukhopadhyay2024diff2lipaudioconditioned}, SadTalker~\cite{zhang2023sadtalkerlearningrealistic}, FLOAT~\cite{ki2025floatgenerativemotion}, Hallo2~\cite{cui2024hallo2longdurationhighresolution}, and LCVD~\cite{guo2025highfidelityrelightablemonocular}. $\uparrow$ indicates higher is better and $\downarrow$ indicates lower is better. \textbf{Bold} and \underline{underlined} values denote the best and second-best results, respectively.}
\label{tab:self}
\end{table*}

\begin{figure*}
\begin{center}
% \fbox{\rule{0pt}{2in} \rule{.9\linewidth}{0pt}}
   \includegraphics[width=0.96\linewidth]{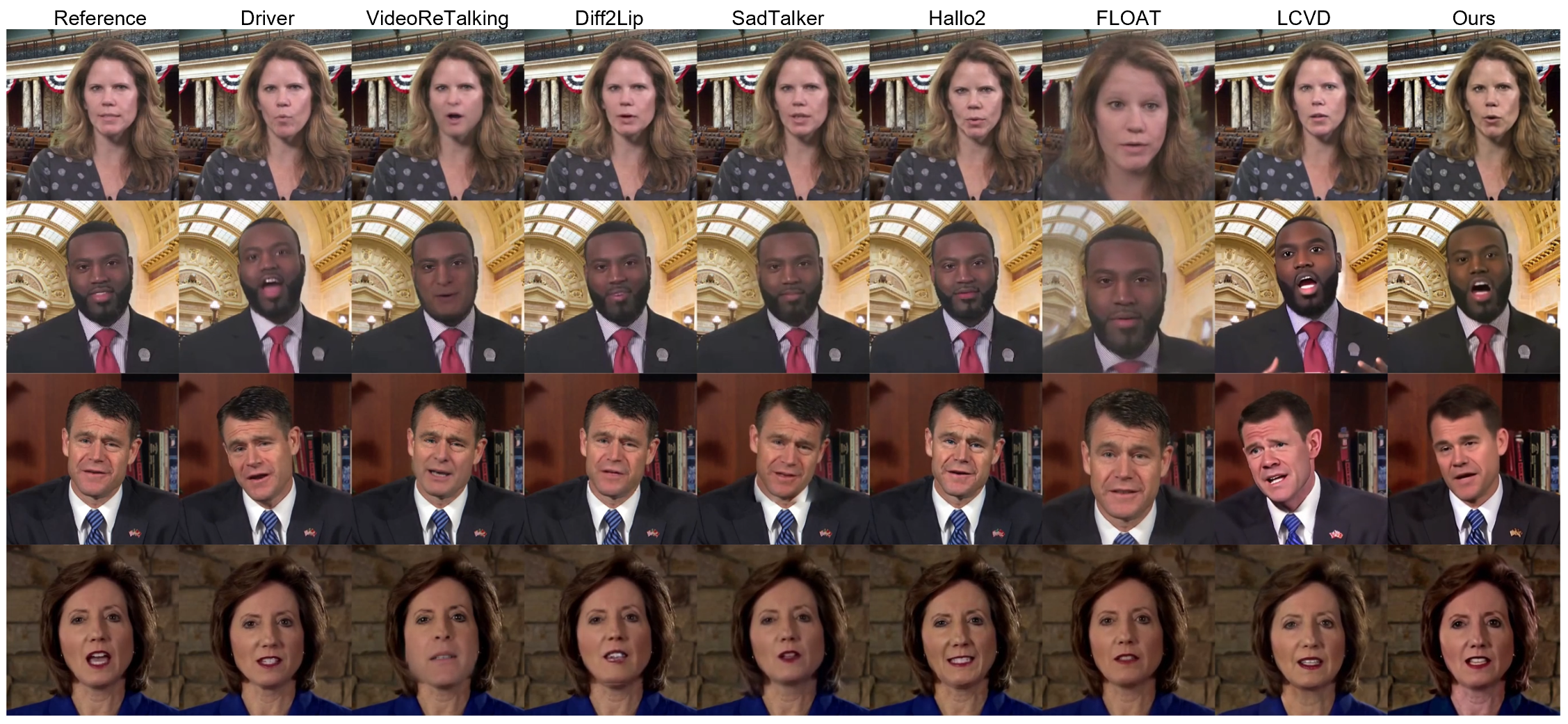}
\end{center}
   \caption{Qualitative comparison of self-reenactment talking-head generation. The first two columns show the reference portrait and the driving sequence, followed by results from VideoReTalking~\cite{cheng2022videoretalkingaudiobasedlip}, Diff2Lip~\cite{mukhopadhyay2024diff2lipaudioconditioned}, SadTalker~\cite{zhang2023sadtalkerlearningrealistic}, FLOAT~\cite{ki2025floatgenerativemotion}, Hallo2~\cite{cui2024hallo2longdurationhighresolution}, LCVD~\cite{guo2025highfidelityrelightablemonocular}, and our approach.}
\label{fig:self}
\end{figure*}

\begin{figure*}
\begin{center}
% \fbox{\rule{0pt}{2in} \rule{.9\linewidth}{0pt}}
   \includegraphics[width=0.96\linewidth]{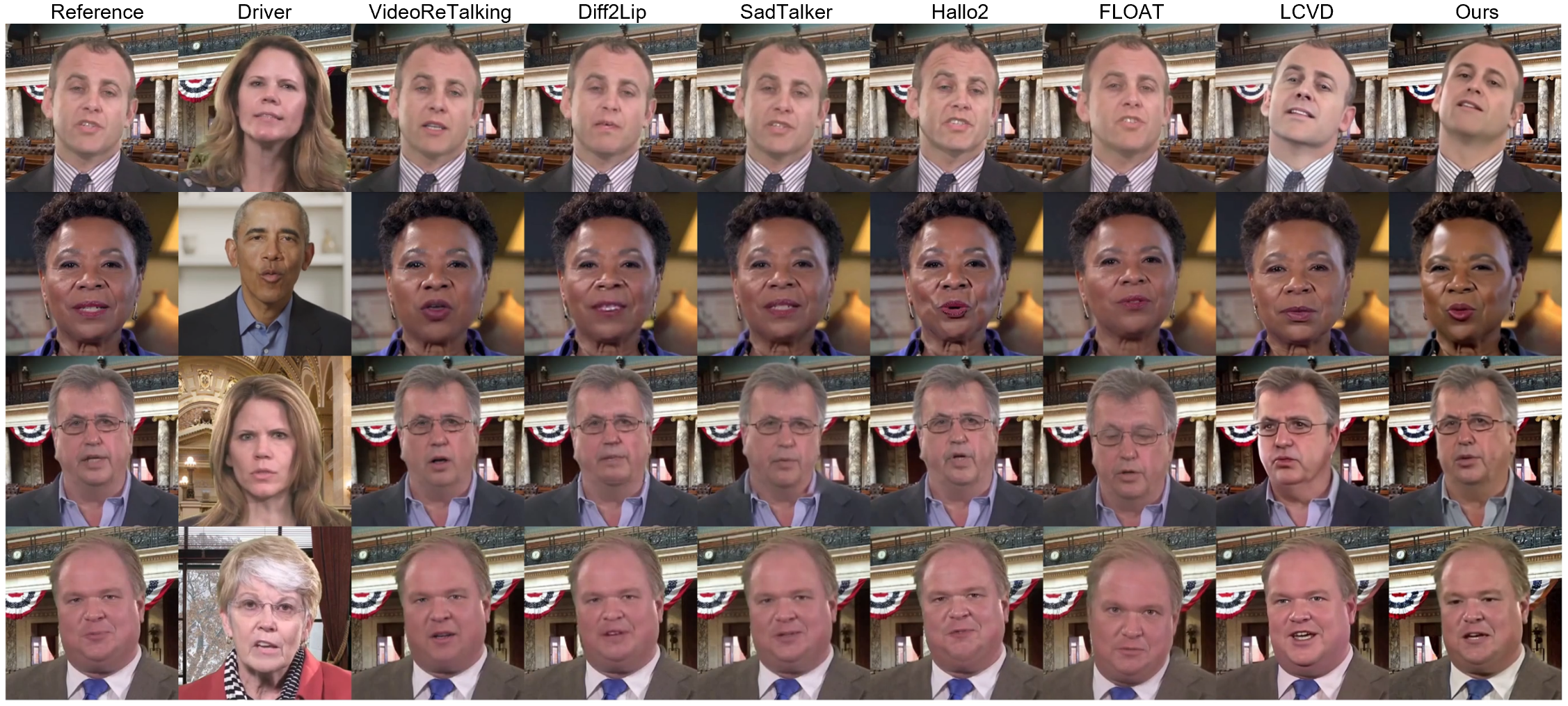}
\end{center}
   \caption{Qualitative comparison of cross-reenactment talking-head generation. The first two columns show the reference portrait and the driving sequence, followed by results from VideoReTalking~\cite{cheng2022videoretalkingaudiobasedlip}, Diff2Lip~\cite{mukhopadhyay2024diff2lipaudioconditioned}, SadTalker~\cite{zhang2023sadtalkerlearningrealistic}, FLOAT~\cite{ki2025floatgenerativemotion}, Hallo2~\cite{cui2024hallo2longdurationhighresolution}, LCVD~\cite{guo2025highfidelityrelightablemonocular}, and our approach.}
\label{fig:cross}
\end{figure*}

% ------------------------------------------------------------
\subsection{Lip Consistency Loss}
\label{sec:method_lip}
We further introduce a lip consistency loss that supervises mouth-region semantics in pixel space. During training, the predicted denoised latents are decoded back to RGB frames, and then a frozen lip-reading encoder~\cite{filntisis2022visualspeechawareperceptuala} extracts mouth-region features
$f_t^{\text{pred}}$ and $f_t^{\text{gt}}$ from each frame pair. The loss is defined as minimizing the cosine distance between them,
\begin{equation}
    \mathcal{L}_{\text{lip}}
        = 1 - \frac{1}{T'} \sum_{t=1}^{T'}
            \cos\!\bigl (f_t^{\text{pred}},\; f_t^{\text{gt}}\bigr),
\end{equation}
where $T'$ is the number of frames used for lip supervision.

\noindent{\bf Training Objective.}
The overall training loss combines the latent denoising objective
$\mathcal{L}_{\text{SVD}}$, an appearance loss
$\mathcal{L}_{\text{app}}$, and a lip consistency loss $\mathcal{L}_{\text{lip}}$,
\begin{equation}
    \mathcal{L}
        = \mathcal{L}_{\text{SVD}}
        + \mathcal{L}_{\text{app}}
        + \mathcal{L}_{\text{lip}}.
\end{equation}

\begin{table*}[t]
\centering
\resizebox{\textwidth}{!}{
\begin{tabular}{|l|c|c|c|c|c|c|c|c|c|c|c|c|c|}
\hline
\multirow{2}{*}{Method} & \multicolumn{10}{c|}{Self-Reenactment} & \multicolumn{3}{c|}{Cross-Reenactment} \\
\cline{2-14}
 & SSIM$\uparrow$ & PSNR$\uparrow$ & LPIPS$\downarrow$ & FVD$\downarrow$ & AED$\downarrow$ & APD$\downarrow$ & AKD$\downarrow$ & VQA$\uparrow$ & ID$\uparrow$ & Sync-C$\uparrow$ & AED$\downarrow$ & APD$\downarrow$ & ID$\uparrow$ \\
\hline\hline
w/o Adaptive Multi-Condition Router & 0.7717 & 23.08 & 0.1286 & 592.5 & 0.1359 & 18.55 & 1.648 & 0.5806 & 0.6739 & 1.157 & 0.2536 & 96.78 & 0.6844 \\
w/o Lip Consistency Loss & 0.7828 & 22.71 & 0.0925 & 93.07 & 0.1117 & 14.84 & \underline{1.456} & \textbf{0.6762} & 0.8043 & 1.171 & \underline{0.2471} & \underline{94.85} & 0.8001 \\
w/o Augmented Mesh & \textbf{0.8004} & \textbf{23.61} & \textbf{0.0805} & \underline{41.51} & \underline{0.1030} & \underline{13.97} & 1.494 & 0.6292 & \underline{0.8380} & \underline{1.182} & 0.2655 & 100.8 & \underline{0.8065} \\
\hline
Ours & \underline{0.7998} & \underline{23.55} & \underline{0.0810} & \textbf{40.75} & \textbf{0.1028} & \textbf{13.85} & \textbf{1.393} & \underline{0.6320} & \textbf{0.8382} & \textbf{1.191} & \textbf{0.2382} & \textbf{89.60} & \textbf{0.8293} \\
\hline
\end{tabular}
}
\vspace{0.005pt} 
\caption{Quantitative ablation evaluation of key components under self-reenactment and cross-reenactment settings. Three key components are validated: Adaptive Multi-Condition Router, Lip Consistency Loss, and Mouth-Augmented Mesh representation. $\uparrow$ indicates higher is better and $\downarrow$ indicates lower is better. \textbf{Bold} and \underline{underlined} values denote the best and second-best results, respectively.}
\label{tab:ablation}
\end{table*}

\section{Experiments}
\subsection{Implementation Details}
\noindent{\bf Datasets.} We train our model on the CelebV-HQ \cite{zhu2022celebvhqlargescalevideo}, MultiTalk \cite{sung-bin2024multitalkenhancing3d}, TFHP \cite{sun2024diffposetalkspeechdrivenstylistic}, MEAD \cite{wang2020meadlargescaleaudiovisual} datasets. Since the backbone of SVD is sensitive to video quality, we first evaluate each video in the datasets with a video quality assessment model \cite{wu2022fastvqaefficientendtoend}, and filtered out videos with scores lower than 0.6. The resulting combined training set contains 34,000 videos.

\noindent{\bf Training Details.} During the training phase, we sample 8-frame video sequences with each frame at a resolution of $512 \times 512$. The model is trained for 30,000 steps with a batch size of 8 using gradient accumulation, optimized by 8-bit-Adam~\cite{kingma2017adammethodstochastic} with a learning rate of $1 \times 10^{-5}$.
To improve robustness and support classifier-free guidance, we apply stochastic condition dropout during training. Each spatial condition branch (reference, motion, shading and audio) is randomly masked with a dropout. This exposes the network to both fully conditioned and partially conditioned scenarios.

\subsection{Evaluation Metrics}
We evaluate under both self- and cross-reenactment settings on
HDTF~\cite{zhang2021flowguidedoneshottalking}, using the first 100
frames of every test video. \emph{Reconstruction fidelity} is measured by
PSNR, SSIM, and LPIPS~\cite{zhang2018unreasonableeffectivenessdeep}.
\emph{Identity preservation} is measured by cosine similarity of
ArcFace embeddings~\cite{deng2019arcfaceadditiveangular}.
\emph{Motion accuracy} is captured by the Average Expression Distance
(AED) and Average Pose Distance (APD) from a 3DMM
estimator~\cite{deng2020accurate3dface}, and the Average Keypoint
Distance (AKD) from a facial landmark
detector~\cite{bulat2017howfarare}. \emph{Perceptual and temporal
quality} is measured by FasterVQA~\cite{wu2022fastvqaefficientendtoend},
the Fréchet Video Distance
(FVD)~\cite{skorokhodov2022styleganvcontinuousvideo}, and the SyncNet
confidence score (Sync-C)~\cite{chung2017outtimeautomated} for lip synchronization. In the cross-reenactment setting, where
ground truth is unavailable, we report AED, APD, and identity similarity, in line with prior work \cite{chu2024generalizableanimatablegaussian,guo2025highfidelityrelightablemonocular}.

\subsection{Quantitative Results}

Table~\ref{tab:self} reports results on HDTF under both reenactment settings. In self-reenactment, our method achieves the best score on seven of the ten metrics, covering all indicators that jointly reflect structural accuracy, motion fidelity, and perceptual reconstruction. The most notable gain is on FVD, where we reduce the second-best score by 37\%, confirming that the generated videos are both temporally coherent and distributionally close to real video. Those results show that our method delivers precise motion transfer while maintaining globally natural dynamics. In cross-reenactment, our method achieves the second-best AED behind FLOAT and the second-best APD behind Hallo2. The consistency of our motion accuracy when the driving identity changes indicates that the learned conditioning is effectively disentangled from appearance, rather than relying on identity-specific cues leaking through the motion signal. 

We note that our method does not lead on ID or Sync-C. Models such as SadTalker and Diff2Lip are specifically engineered around a single objective and enforce it through task-specific losses. In contrast, our method prioritizes controllability across multiple complementary attributes, so that expression, head motion, mouth dynamics, and lighting can be independently specified at inference.

\subsection{Qualitative Results}

Figures~\ref{fig:self} and~\ref{fig:cross} present visual comparisons of our method against six baselines. Diff2Lip generates
limited head motion and produces visible artefacts in facial expressions and mouth regions. SadTalker preserves identity
reasonably well but exhibits a discontinuity at the face-neck
boundary, as if the animated facial region were overlaid onto a static neck. Hallo2 and FLOAT achieve audio-aligned lip movements and generate natural facial expressions and head motions, but they lack fine-grained controllability over individual facial attributes, limiting their applicability in customizable animation scenarios. LCVD achieves accurate motion transfer, but suffers
from two shortcomings: a degradation in the fidelity of facial features, and a frame-level stuttering that stems from its reliance on a per-frame image-to-3DMM reconstruction without any temporal prior. In contrast, our method achieves a balanced performance across all these axes: it preserves reference appearance, accurately transfers the specified motion, and maintains smooth temporal continuity between frames.

\subsection{Ablation Study}
In Table~\ref{tab:ablation}, we ablate the three key components of our framework: Adaptive Multi-Condition Router, Lip Consistency Loss, and Mouth-Augmented Mesh
representation.

\noindent{\bf Effect of the Adaptive Multi-Condition Router.}
Removing the router and reverting to naive additive fusion of the four condition streams causes a drastic collapse across nearly all metrics. Under self-reenactment, FVD jumps from $40.75$ to
$592.5$ (a $14\times$ degradation), and ID drops by $19.6\%$. The damage is similarly severe under cross-reenactment: ID falls by $17.5\%$, and APD rises from 89.60 to 96.78. The simultaneous collapse of distributional, identity, and motion metrics across both settings indicates that without adaptive gating, the four heterogeneous conditioning signals interfere destructively within the U-Net.

\noindent{\bf Effect of the Lip Consistency Loss.}
Disabling the lip consistency loss preserves overall reconstruction quality but causes a small yet consistent drop in motion-related and synchronization metrics. Interestingly, this variant achieves the best VQA, suggesting that without the mouth-region constraint the model produces marginally smoother per-frame textures at the cost of fine-grained lip dynamics.

\noindent{\bf Effect of the Mouth-Augmented Mesh.}
The variant without the mouth-augmented mesh achieves marginally better single-frame reconstruction metrics (SSIM, PSNR, LPIPS) than the full model. However, this advantage disappears once temporal coherence and identity disentanglement are taken into account. The full model consistently outperforms this variant on temporal and motion-related metrics under self-reenactment, and the gap widens sharply under cross-reenactment, with clear degradation across AED, APD, and ID. We attribute this asymmetry to the insensitivity of frame-level metrics to temporal instability: shared identity in self-reenactment partially masks the weakness of an unstable geometric prior. When the driving identity differs, the absence of temporally coherent geometry exposes the jitter inherent in DECA's single-image fitting, significantly degrading motion fidelity.

Taken together, the ablations validate a clear hierarchy of contributions. The Adaptive Multi-Condition Router is indispensable for multi-conditional fusion and is the dominant driver of overall quality. The Mouth-Augmented Mesh trades a negligible frame-level cost for a measurable gain in temporal coherence and audio-visual synchronization, serving as a temporally stable geometric prior for identity-agnostic motion transfer. The Lip Consistency Loss provides a small but targeted refinement of mouth dynamics. 
% The fact that each component yields larger relative gains under cross-reenactment than under self-reenactment highlights the robustness of our design, particularly its ability to maintain precise motion control without leaking identity cues.

\section{Limitation and Future Work}

Our approach has two main limitations. First, while the framework achieves competitive overall quality, there remains room for improvement in identity preservation and lip-sync accuracy. Second, the backbone generates only 8 frames per forward pass, and longer sequences are produced by stitching consecutive segments with temporal overlap, which inherently limits scalability and may introduce subtle discontinuities at segment boundaries. Future work will explore stronger constraints for identity preservation and lip synchronization, and migrate to an autoregressive backbone~\cite{gao2025dardiffusionautoregressive} to enable continuous, arbitrarily long generation within a single pass.

\section{Conclusion}
We have presented a multi-conditional video diffusion framework for controllable talking-head generation that unifies audio, keypoints, 3DMM mesh, and reference appearance. At its core, the Adaptive Multi-Condition Router computes channel-wise weighting over the four condition streams, allowing the fusion strategy to vary with both feature subspace and noise level. The Mouth-Augmented Mesh combines DECA's identity and lighting priors with SPECTRE's mouth tracking to provide a temporally coherent geometric prior for the lower face, while a Lip Consistency Loss further tightens audio-visual alignment. Extensive experiments confirm that our framework delivers strong reconstruction, motion fidelity, and temporal coherence, while providing robust attribute-level controllability that single-condition baselines do not offer.

{\small
\bibliographystyle{ieee}
\bibliography{MoCoTalk}
}

\appendix
\section{Implementation Details}
 
% ------------------------------------------------------------
\subsection{Lip Consistency Loss}
\label{sec:appendix_lip}
While the latent denoising objective enforces global reconstruction 
fidelity, it provides only weak supervision for fine-grained mouth 
dynamics. To address this, we introduce a \emph{lip consistency loss} 
that supervises mouth-region semantics directly in pixel space using 
a frozen visual speech recognition encoder.

\noindent{\bf Mouth region extraction.}
During training, the predicted denoised latents are decoded back to RGB frames and resized to $224 \times 224$ to match the input resolution of the lip-reading encoder. We then detect facial landmarks on the ground-truth frames and derive a single, temporally stable mouth bounding box by aggregating tracked landmarks across the entire clip, rather than re-estimating the crop independently for each frame. The same bounding box is then applied to both the predicted and ground-truth frames. This design avoids per-frame jitter in the crop region, which would otherwise inject spurious motion into the supervision signal and confound genuine lip articulation with cropping noise. We denote this mouth cropping and normalization operator by $\mathcal{M}(\cdot)$.

\noindent{\bf Feature extraction and loss.}
Let $\mathcal{E}_{\text{lip}}(\cdot)$ denote the frozen lip-reading encoder~\cite{filntisis2022visualspeechawareperceptuala}, which has been pretrained on large-scale visual speech recognition and is therefore sensitive to phoneme-level mouth dynamics. Given the ground-truth video $\mathbf{X}$ and the decoded prediction $\hat{\mathbf{X}}$, we extract per-frame mouth features
\begin{equation}
    \mathbf{f}^{\text{gt}} = \mathcal{E}_{\text{lip}}\bigl(\mathcal{M}(\mathbf{X})\bigr),
    \qquad
    \mathbf{f}^{\text{pred}} = \mathcal{E}_{\text{lip}}\bigl(\mathcal{M}(\hat{\mathbf{X}})\bigr).
\end{equation}
The lip consistency loss is defined as in Sec.~\ref{sec:method_lip} of the main 
paper. Because decoding latents back to RGB and propagating gradients through the lip-reading encoder is memory-intensive, we sample only a subset of $T' = 2$ frames per training step in practice, which we find sufficient to provide a stable supervisory signal without inflating GPU memory consumption.

\subsection{Training Details}
Training is performed on a single NVIDIA RTX 6000 GPU (32GB) and takes 
approximately 7 days. At inference, generating a 100-frame video on 
the same GPU takes approximately 2 minutes.
\end{document}